\documentclass[runningheads]{llncs}
\usepackage{graphicx}
\usepackage{comment}
\usepackage{amsmath,amssymb} 
\usepackage{color}
\usepackage{mathrsfs}
\usepackage{algorithm}  
\usepackage[noend]{algpseudocode}
\usepackage{algorithmicx,algorithm}
\usepackage{threeparttable}
\usepackage{caption}
\usepackage{subcaption}
\newcommand\blfootnote[1]{%
	\begingroup
	\renewcommand\thefootnote{}\footnote{#1}%
	\addtocounter{footnote}{-1}%
	\endgroup
}

\newcommand{\etal}{\textit{et al}. }
\newcommand{\ie}{\textit{i}.\textit{e}. }

\usepackage{multirow}

\begin{document}
	\captionsetup[figure]{labelsep=period}
	\captionsetup[table]{labelsep=period}
	\pagestyle{headings}
	\mainmatter
	\def\ECCVSubNumber{386}  
	
	\title{Differentiable Hierarchical Graph Grouping for Multi-Person Pose Estimation}

	\titlerunning{Differentiable Hierarchical Graph Grouping}
	%
	\author{Sheng Jin\inst{1,2}\orcidID{0000-0001-5736-7434} \and
		Wentao Liu\inst{2\dagger}\orcidID{0000-0001-6587-9878} \and \\
		Enze Xie\inst{1} \and Wenhai Wang\inst{3} \and
		Chen Qian\inst{2} \and Wanli Ouyang\inst{4} \and Ping Luo\inst{1}  \\[.21cm]
		$^{1}$ The University of Hong Kong \quad
		$^{2}$ SenseTime Research \\
		$^{3}$ Nanjing University \quad
		$^{4}$ The University of Sydney \\
		\tt\small \{jinsheng, liuwentao, qianchen\}@sensetime.com \quad
		wanli.ouyang@sydney.edu.au, pluo@cs.hku.hk}
	
	\authorrunning{S. Jin et al.}
	%
	\institute{}
	\maketitle
	
	\blfootnote{$^{\dagger}$Corresponding author.}
	\begin{abstract}
		Multi-person pose estimation is challenging because it localizes body keypoints for multiple persons simultaneously. Previous methods can be divided into two streams, \ie top-down and bottom-up methods. The top-down methods localize keypoints after human detection, while the bottom-up methods localize keypoints directly and then cluster/group them for different persons, which are generally more efficient than top-down methods. However, in existing bottom-up methods, the keypoint grouping is usually solved independently from keypoint detection, making them not end-to-end trainable and have sub-optimal performance. In this paper, we investigate a new perspective of human part grouping and reformulate it as a graph clustering task. Especially, we propose a novel differentiable Hierarchical Graph Grouping (HGG) method to learn the graph grouping in bottom-up multi-person pose estimation task. Moreover, HGG is easily embedded into main-stream bottom-up methods. It takes human keypoint candidates as graph nodes and clusters keypoints in a multi-layer graph neural network model. The modules of HGG can be trained end-to-end with the keypoint detection network and is able to supervise the grouping process in a hierarchical manner. To improve the discrimination of the clustering, we add a set of edge discriminators and macro-node discriminators. Extensive experiments on both COCO and OCHuman datasets demonstrate that the proposed method improves the performance of bottom-up pose estimation methods.
		\keywords{Human Pose Estimation, Graph Neural Network, Grouping}
	\end{abstract}
	
	\section{Introduction}
	
	\begin{figure}
		\begin{center}
			\includegraphics[width=0.95\linewidth]{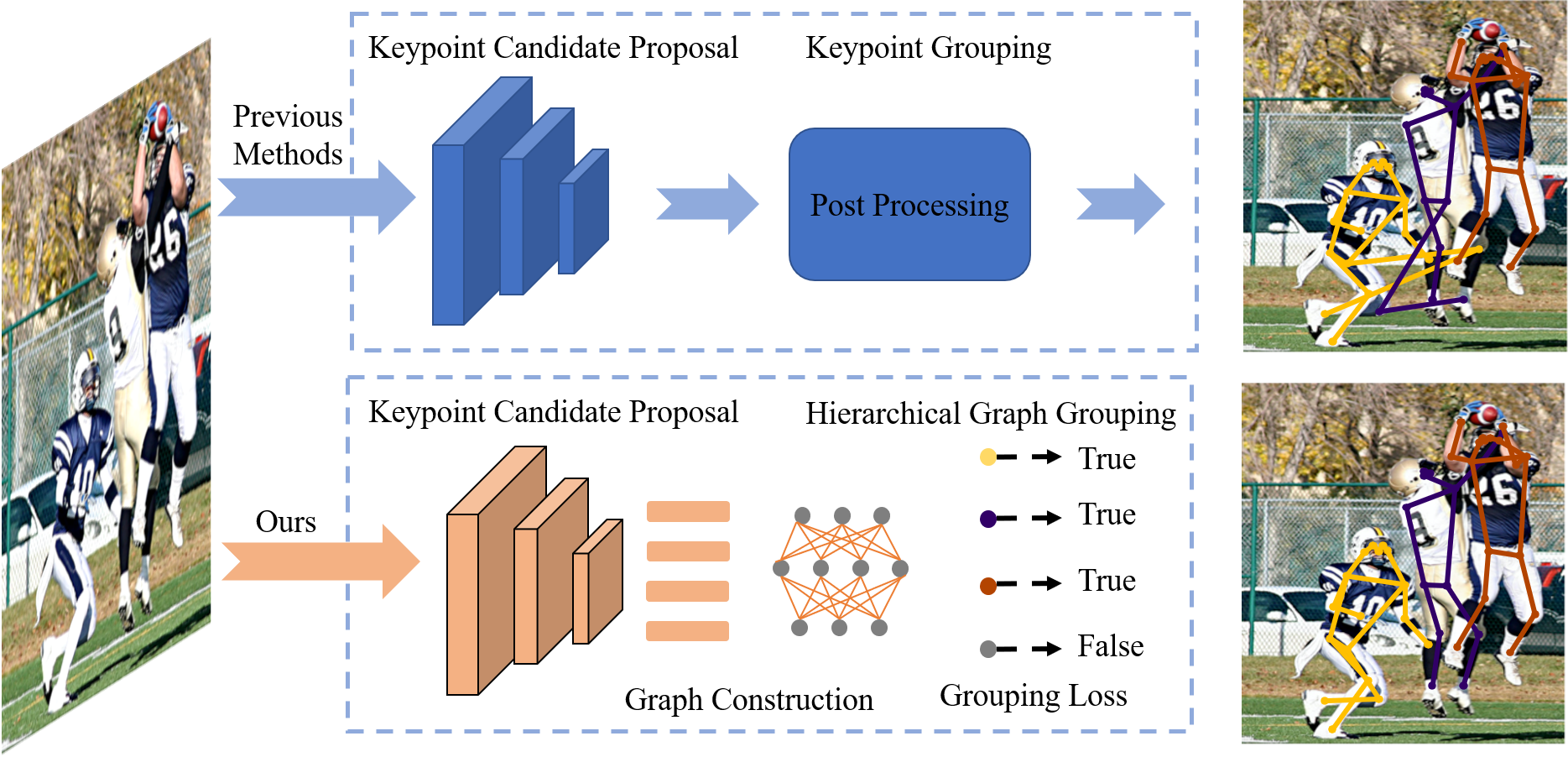}
		\end{center}
		\caption{ Hierarchical Graph Grouping embeds grouping procedure with the keypoint candidate proposal network. All modules are differentiable and can be trained end-to-end. Keypoint candidates are grouped in a multi-layer graph neural network, which enables to directly supervise the final grouping results. }
		\label{fig:motivation}
	\end{figure}
	
	Multi-person pose estimation aims at localizing 2d keypoints of an unknown number of people in an image. It has attracted much research interest because of its significance in various real-world applications, such as human behavior understanding, human-computer interaction, and action recognition.
	
	Current pose estimation methods perform keypoints detection in two routes. The \emph{top-down} methods  \cite{chen2018cascaded,he2017mask,li2019crowdpose,papandreou2017towards,sun2019deep,sun2018integral,xiao2018simple} first detect human bounding boxes and then estimate keypoints for each person. It performs a single person pose estimation to all human candidates, so it is often time-consuming.
	Contrarily, \emph{bottom-up} pose estimation approaches~\cite{cao2017realtime,jin2019multi,newell2017associative,papandreou2018personlab} follow the keypoints detection-and-grouping pipeline: detecting keypoints at the first stage and grouping them into individuals at the second stage. These methods are more efficient and have gained increasing attention in the industry. 
	Previous works generally treat the grouping stage as post-processing by using integer linear programming~\cite{Insafutdinov2016ArtTrack,Insafutdinov2016DeeperCut,jin2017towards,pishchulin2016deepcut}, heuristic greedy parsing~\cite{cao2017realtime,papandreou2018personlab}, or clustering~\cite{newell2017associative,nie2017generative}. But they are not able to be trained end-to-end, which is in conflict with deep learning’s philosophy of learning everything together. Previous bottom-up methods generally learn some substitute indicators which may reflect the grouping accuracy, resulting in sub-optimal solutions. 
	For example, associate embedding (AE) ~\cite{newell2017associative} produces the permutation-invariant associative embedding (a vector representation) for each keypoint, and learns by pushing apart the embedding of different people and pulling closer that of the same instance. Although it uses the associative embedding which encodes pairwise relationship to group keypoints, the grouping procedure itself is still offline, and no direct supervision is applied to the grouping results. There is a mismatch between the pairwise loss and the accuracy of the greedy parsing used at inference time. Even though the pairwise loss is low, the parsing results can still be possibly wrong, and vice versa. 
	
	A better choice is to directly supervise the grouping process. 
	However, one major challenge is that the previous keypoint grouping procedure is often not differentiable, and thus is hard to be integrated with keypoint detection. Moreover, how to deal with the flexible number of keypoints is still an open problem.
	
	In this paper, we present a simple and elegant solution for bottom-up multi-person pose estimation. In the proposed method, 
	the whole network, composed of a keypoint detection network and a grouping network, is \emph{fully end-to-end trainable},
	and able to	flexibly deal with the grouping problem of a variable number of human instances.
	To achieve this, we first reformulate the grouping problem as the graph clustering problem. 
	A graph corresponds to an image, where the nodes denote the keypoint proposals, and edges denote whether the two keypoints belong to the same person. The graph structure is adaptive to different input images instead of constructing a static graph, so it is able to dynamically group various numbers of keypoints into various numbers of human instances. Especially, we propose the Online Hierarchical Graph Clustering (OHGC) algorithm,
	which makes the process of grouping keypoints learnable and can be easily embedded into main-stream bottom-up methods. The HGG method initializes the graph from the keypoint proposal network and groups pairs of most relative nodes in each iteration through the OHGC algorithm.
	
	In OHGC, keypoints are clustered step-by-step.
	Each keypoint proposal starts in its own graph node, and the cluster pairs are merged. This forms a pose hierarchy, from small fractions to the whole body. This enables the model to pay more attention to global consistency and learn effective features for predicting the pairwise relation.
	The group operations are fully differentiable, so OHGC can make the whole network~(including keypoint detection and grouping) end-to-end trainable. By directly supervising the grouping results, the grouping loss is back-propagated to the previous keypoint detection network, which will further improve the feature representation ability of the keypoint detection network. 
	
	Moreover, we propose the edge discriminator to strengthen the local relationship of keypoints, and the macro-node discriminator to enforce global consistency. It can further increase the discrimination of body-keypoint relational features, leading to better grouping accuracy.
	
	The main contributions of this work are thus three-fold.	
	\begin{itemize}
		\itemsep 0.1cm
		\item We reformulate the task of multi-person pose estimation as a graph clustering problem and present the first fully end-to-end trainable framework with grouping supervision for bottom-up multi-person pose estimation.
		\item We propose edge discriminators and macro-node discriminators to learn both local and global pairwise relation features and boost the grouping accuracy. 
		\item The experimental results show that the proposed method outperforms the baseline by a large margin and achieves comparable performance with the state-of-the-art bottom-up pose estimation methods on COCO dataset. Moreover, the proposed method achieves the state of the art performance on the OCHuman datasets (41.8/36.0 mAP for val and test respectively).
	\end{itemize}
	
	\section{Related Work}
	\label{sec:related_work}
	
	\subsection{Multi-person Pose Estimation in Images}
	\textbf{Top-down} methods~\cite{chen2018cascaded,fang2017rmpe,he2017mask,huang2017coarse,li2019crowdpose,liu2018cascaded,papandreou2017towards,sun2019deep,xiao2018simple} decompose the multi-person pose estimation task into two sub-tasks:(1) Human detection and (2) Pose Estimation in the region of a single human. 
	First, the person detector predicts a bounding box for every human instance in the image. Second, the box is cropped and resized from the image. Third, single-person pose estimation is applied to predict the keypoints for the cropped person. 
	In addition, some work such as Mask R-CNN~\cite{he2017mask} crop the feature instead of raw images to boost efficiency. 
	In summary, top-down methods are dominant in state-of-the-art methods but they often have higher computational complexity overhead, especially when the number of human instances increases. This is because they need to repeatedly run the single-person pose estimation for every instance. Furthermore, because the pose estimation is dependent on the detection, it is difficult for these methods to recover the pose of an instance if it is missing in the detection results.
	
	\textbf{Bottom-up} approaches~\cite{cao2017realtime,Insafutdinov2016ArtTrack,Insafutdinov2016DeeperCut,Iqbal2016PoseTrack,jin2019multi,jin2017towards,newell2017associative,nie2017generative,papandreou2018personlab,pishchulin2016deepcut} first detect all keypoint candidates in an image, then assemble/group them into full-body keypoints of each instance. Such bottom-up methods are usually efficient, and are capable of achieving real-time performance. To aid the follow-up keypoint association, most bottom-up methods learn descriptors to encode keypoint pairwise relations and to distinguish different instances. PAF~\cite{cao2017realtime} learns part-affinity-fields, encoding both the location and orientation of keypoint pairs; GPN~\cite{nie2017generative} learns 2D offset fields, linking keypoints to the corresponding human centers; PersonLab~\cite{papandreou2018personlab} introduces long-range, mid-range and short-range offsets between pairwise keypoints; AE~\cite{newell2017associative} learns the associative embedding for each keypoint and similar embedding indicates higher possibility of belonging to the same person. The grouping process is generally formulated as a post-processing optimization problem and solved by graph partitioning~\cite{Insafutdinov2016ArtTrack,Insafutdinov2016DeeperCut,Iqbal2016PoseTrack,pishchulin2016deepcut}, heuristic greedy decoding algorithm~\cite{cao2017realtime,papandreou2018personlab} or spectral clustering~\cite{nie2017generative}.
	In summary, bottom-up methods can benefit from sharing convolutional computation, as a result, being faster than top-down methods. Nevertheless, the post-processing of grouping is heuristic and involves many hyper-parameters. Since the pose estimation and post-processing are not jointly learned, they cannot collaborate and adapt to each other.
	Instead of regarding the grouping as a pure post-processing procedure, we propose to train grouping with pose estimation jointly in an end-to-end fashion, enabling the error signals for grouping to be back-propagated.
	
	\textbf{Single-stage pose estimation.}
	With recent advantages of single-shot object detection and instance segmentation~\cite{tian2019fcos,xie2019polarmask,zhou2019objects}, some single-stage pose estimation methods are proposed. CenterNet~\cite{zhou2019objects} firstly transfer pose estimation as human center detection and keypoint regression. However, it still needs keypoint detection and projection to improve performance.
	SPM~\cite{nie2019single} proposes a structured pose representation to divide the keypoints hierarchically. In this way, it can ease the difficulty of long-range regression. 
	Similarly, DirectPose~\cite{tian2019directpose}, based on FCOS~\cite{tian2019fcos}, directly do human center classification and keypoint regression without relying on bounding box. KPAlign is proposed to overcome the feature misalignment between convolutional features and keypoint predictions.
	However, keypoint regression is not very precise in the above methods, especially under the restriction of High IoU. In comparison, our method retains higher precision, especially under more strict metrics (AP$_{75}$).
	
	\subsection{Graph Representation for Pose Estimation}
	
	The graph representation for human pose estimation is not new. For single-person pose estimation, many work~\cite{chen2014detect,chen2014articulated,chu2016crf,felzenszwalb2005pictorial,fischler1973representation,johnson2010clustered,tompson2014joint,yang2012articulated} have been based on various graphical models such as pictorial structure, Mixtures-of-parts, Markov Random Fields (MRF) or Conditional Random Fields (CRF). In these works, the graph nodes represent keypoints and the edges encode the pairwise relationships between keypoints. Since all the keypoints belong to the same human instance, no grouping process is required. Moreover, the number of keypoints of a single person is always fixed, therefore the graph structure, in terms of the number of nodes and the connectivity of edges, is fixed. 
	
	Multi-person pose estimation is much more challenging. ~\cite{Insafutdinov2016ArtTrack,Iqbal2016PoseTrack,wang2018bi}, the pose estimation problem is cast as a graph partitioning based integer linear programming (ILP) problem. However, the optimization process is offline and very time-consuming. Song \etal~\cite{song2019end} proposed a method for end-to-end minimum cost multicut problem. Unlike their works which focus on the CRF optimization, we solve the keypoint grouping task by direct graph clustering.
	
	\subsection{Graph Neural Networks}
	
	This paper reformulates the multi-person pose estimation task using the graph representation and applies graph neural networks to this problem. Graph Neural Networks (GNN) is initially introduced in~\cite{gori2005new,scarselli2008graph} and has become a popular tool for efficient message passing and modeling global relations~\cite{chen2019graph}. Most of GNN models can be categorized into two types: spectral approaches~\cite{bruna2013spectral,kipf2016semi} and non-spectral approaches~\cite{duvenaud2015convolutional,wang2019dynamic}. This work is related to~\cite{wang2019dynamic}, which efficiently models the edge features. To solve the task of multi-person pose estimation, based on~\cite{wang2019dynamic} we develop a hierarchical clustering method, which takes the body structure constraints into consideration and models the whole grouping process.
	
	More recently, GNN models have been applied to model the human body structure. Yan~\etal~\cite{yan2018spatial} proposes the spatial-temporal graph convolutional networks for skeleton-based action recognition. Zhang \etal~\cite{zhang2019human} proposes to use PGNN to learn the structured representation of keypoints for single-person pose estimation. However, previous works only deal with the single person case, where the structure of the graph is fixed. The multi-person case is more challenging, since the number of keypoints and the number of people vary in different images and even in different grouping stages. We have to develop a dynamic graph interaction model to effectively handle such problems.
	
	\section{Method}
	
	\subsubsection{Overview}
\begin{figure*}[t]
		\begin{center}
			\includegraphics[width=0.95\linewidth]{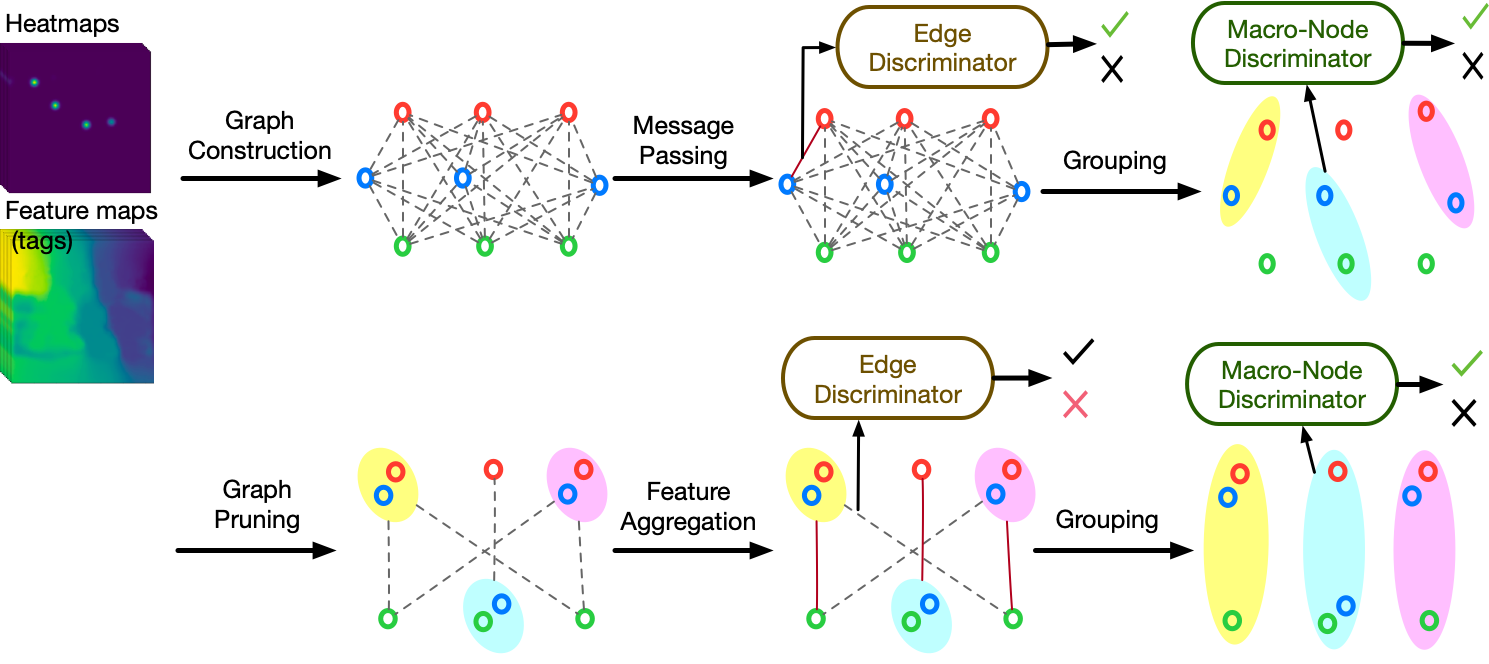}
		\end{center}
		\caption{The keypoint grouping stage of HGG framework. We construct a graph on top of the keypoint candidate proposal network, perform message passing with GNNs, and group the candidates iteratively. Edge discriminators and macro-node discriminators are applied to improve the grouping performance.}
		\label{fig:overall_framework}
	\end{figure*}

	An overview of our proposed hierarchical graph grouping (HGG) framework is illustrated in Fig~\ref{fig:overall_framework}. Our HGG framework consists of two stages, \ie the keypoint candidate proposal stage and the keypoint grouping stage. 
	
	In the keypoint candidate proposal stage, all keypoint candidates are detected and corresponding feature maps are extracted. Following AE~\cite{newell2017associative}, we use a 4-stacked hourglass~\cite{newell2016stacked} as the backbone of the keypoint candidate proposal network. The keypoint proposal network then provides keypoint candidates and raw relational feature embedding for the keypoints grouping module.
	
	In the keypoint grouping stage, we build a graph neural network using the candidates and relational features extracted from the former stage. An online hierarchical graph clustering (OHGC) algorithm is devised to cluster keypoints iteratively. In each iteration, OHGC updates the pairwise relation features and clusters nodes into a \textit{macro-node} by maximizing the weighted edge score. The graph is updated and pruned with respect to the macro-nodes. Contrary to integer linear programming or bipartite matching, the proposed method is fully differentiable and is able to be trained end-to-end with keypoint detection.
	
	We proposed two kinds of the discriminator to strengthen the grouping procedure, the edge discriminator and the macro-node discriminator. In each iteration, the edge discriminator is introduced to classify whether the pair of nodes belong to the same person. The pairwise relation features and the edge scores are updated accordingly. After each iteration of grouping, a macro-node discriminator is applied to each cluster to discriminate between a correctly-clustered macro-node (in which all nodes belong to the same person) and a wrongly-clustered one. In this way, the whole online grouping procedure is fully supervised.
	
	\subsection{Hierarchical Graph Grouping}
	
	Previous work~\cite{Insafutdinov2016ArtTrack,Insafutdinov2016DeeperCut,Iqbal2016PoseTrack,pishchulin2016deepcut} cast the problem of multi-person pose grouping as graph partitioning, and solve it by optimizing an integer linear programming (ILP) problem. However, the optimization process is performed offline and the grouping procedure is not able to be supervised with the keypoint candidate proposal network. In this paper, we rethink this problem from the perspective of graph clustering and solve it with supervised learning. We follow the online agglomerative graph clustering setting. Each keypoint candidate starts with being its own cluster and closest pairs of clusters are merged iteratively. As a result, the keypoint candidates are grouped into several clusters, where each cluster contains all the keypoints of a single person. We are able to directly supervise the final grouping results. In the following sections, we will give a detailed description of the graph construction and hierarchical graph grouping. 
	
	\subsubsection{Graph Construction}
	We construct a graph on top of the keypoint candidate proposal network. In the graph  $\mathbf{G} = \{\mathbf{V}, \mathbf{E}\}$, the ``vertices'' $\mathbf{\{V\}} = \{v_{i}\}_{i=1:N}$ represent keypoint candidates and the ``edges'' $\mathbf{\{E\}} = \{e_{i_1,i_2}\}_{i_1=1:N, i_2=1:N}$ represent the pairwise relationship between the two candidates (the possibility of belonging to the same person or not). Note that the graph is constructed dynamically, as the graph may have different number of nodes and edges for different images. We choose the fully-connected graph that densely connects every pair of the keypoints with different keypoint types. The keypoints with the same type (both ``head''s) are disconnected. Compared to other sparse graph configurations (such as the tree-structure), the fully-connected graph is able to avoid over-segment of a person during occlusion, \ie dividing a single pose into several clusters. For example, when a person's torso is occluded or missing, the link between the head and the foot will be helpful to connect the upper and the lower parts. Moreover, since the number of keypoints in an image is only about 30 on average, the computational cost of constructing such a dense graph is almost negligible. Each vertex $v_i \in \mathbf{\{V\}}$ in the graph is initialized with the concatenation of the following features: (1) the relational embedding features of the keypoint, (2) the one hot feature that encodes the keypoint type, (3) the (x, y) coordinates of the keypoint normalized to $[0, 1]$. Both visual features and spatial features are preserved.
	
	\subsubsection{Online Hierarchical Graph Clustering Algorithm}
	OHGC algorithm is given in Algorithm~\ref{alg:OHGC}. Given the initial graph, an \emph{interaction GNN} (Graph Neural Network) is trained to extract the relational features via message passing between vertices. As shown in Fig.~\ref{fig:network}, our GNN utilizes a stack of EdgeConv~\cite{wang2019dynamic} layers for effective feature learning. In each EdgeConv layer, the edge feature is mapped from the concatenation of features of nodes (linked by the edge) using a fully-connected layer, and the node features are updated by aggregating the features of the associated edges. A three-layer MLP (Multi-layer Perceptron) with Dropout is adopted to further extract high-level node features. As the output, we get representative features of each of the vertex which is used for grouping.
	
	Previous graph clustering algorithms mainly focus on the keypoint-level pairwise relationship, without considering the higher-order term, \ie the relation between two clusters of body parts. We instead propose to model the whole grouping process and design a hierarchical graph clustering algorithm. OHGC repeatedly performs graph feature aggregation, edge proximity update, node clustering and graph pruning, until all the edges are cut. 
	
	In each iteration, \emph{feature aggregation} is applied to each of the macro-node (the set of previously grouped nodes) by averaging all features in the set. The proximity score between macro-nodes is measured by the edge discriminator (see Sec.\ref{sec:discriminators}). After updating the edge weights, we use \emph{graclus clustering}~\cite{dhillon2007weighted} to match each vertex with its neighbors by (approximately) maximizing the edge weights. This finds the most confident pairs and carries out the clustering action. As a result, a group of ``low-level'' nodes is clustered into a ``higher-level'' macro-node. The number of clusters is reduced by half. For COCO dataset, the number of keypoint types is $J=17$, so the grouping will stop in no more than $\lceil \log_2 17 \rceil = 5$ iterations. After that, a macro-node discriminator (see Sec.\ref{sec:discriminators}) is applied to each cluster to discriminate between a correctly-clustered macro-node (in which all nodes belong to the same person) and a wrongly-clustered one. The grouping procedure should satisfy the following two constraints. 1) A keypoint cannot be assigned to more than one person, \ie two people share a single ``head'' keypoint. 2) A person cannot have more than one keypoints of the same type, \ie a person containing two ``head'' keypoints. To avoid infeasible clustering, we perform \emph{graph pruning} to remove infeasible edges after each grouping iteration. If two (macro-)nodes contain the same type of nodes, the edge in between is pruned. This grouping procedure repeats until all edges are pruned. 
	
	This grouping procedure naturally forms a hierarchy, from isolated keypoints to a whole body. The model learns to first group easy-to-group parts, then perform cluster in the macro-node level. As the grouping continues, the graph gradually gets coarsened. Finally, the nodes will be clustered into $K$ groups, indicating $K$ human instances. The model learns to group from easy to hard, in a curriculum fashion~\cite{bengio2009curriculum}. Unlike the previous curriculum learning paradigm which requires to manually set curriculum phases, our curriculum tasks are automatically generated during training and well adjusted to the model’s current capability. 
	
	\begin{algorithm}
		\caption{Online Hierarchical Graph Clustering}
		\hspace*{0.02in} {\bf Input:}
		An RGB image; \\
		\hspace*{0.02in} {\bf Output:} 
		Body pose clusters;
		\begin{algorithmic}
			\State Keypoint candidate proposals;
			\State Graph construction: $\mathbf{G} = \{\mathbf{V}, \mathbf{E}\}$;
			\State Relational feature learning with interaction GNN.
			\Repeat 
			\State Feature aggregation via avg-pooling;
			\State Update the proximity between (macro-)nodes;
			\State Apply graclus clustering; 
			\State Graph pruning;
			\Until{No edges are remained.} 
		\end{algorithmic}
		\label{alg:OHGC}
	\end{algorithm}
	
	\subsection{Grouping Discriminators}
	\label{sec:discriminators}
	In OHGC, two types of discriminators are introduced to further improve the grouping performance. In each iteration, we utilize the edge discriminator to update proximity scores and the macro-node discriminator to suppress the incorrectly grouped macro-nodes. We use the same discriminators in each clustering loop iteration. The network architectures are demonstrated in Fig.~\ref{fig:network}. Binary cross-entropy (BCE) loss is used to train.
	
	\subsubsection{Edge Discriminator.} 
\begin{figure*}[t]
		\begin{center}
			\includegraphics[width=0.99\linewidth]{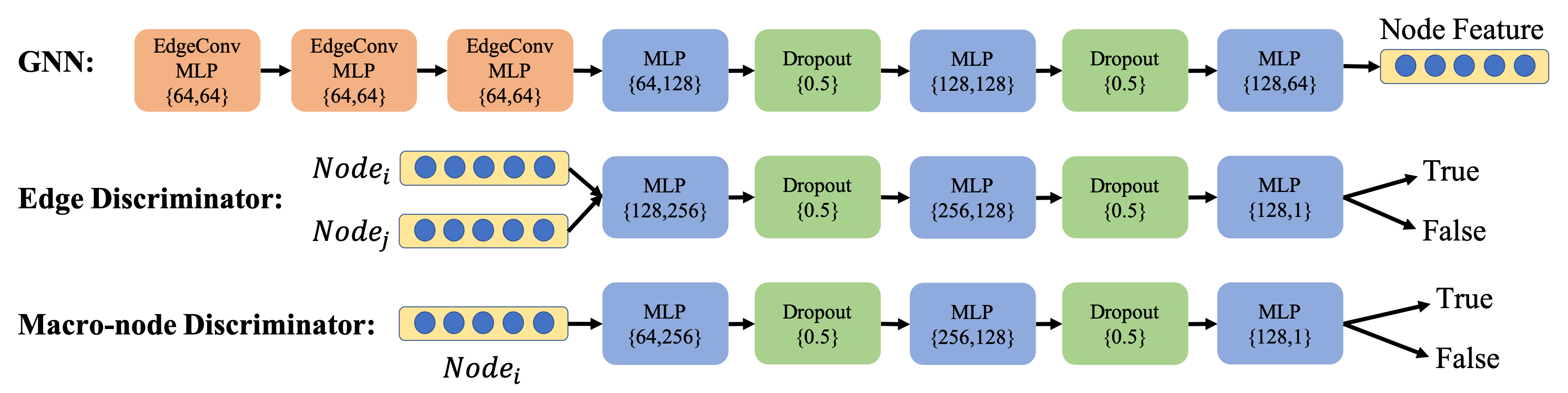}
		\end{center}
		\caption{The network architecture of GNN, the edge discriminator and the macro-edge discriminator. The number of the input/output channels of MLP are given.}
		\label{fig:network}
	\end{figure*}

	Edges preserve local but discriminate keypoint-to-keypoint relationship. In order to improve the discrimination ability of the pairwise relation feature, we introduce a \emph{shared} edge discriminator at each iteration. The edge discriminator is a two-class discriminator that is used to directly classifying the states of the edges: whether the edge is connected (label 1) or not (label 0). Connected edge means the two keypoints belong to the same person. As shown in Fig.~\ref{fig:network}, the edge discriminator is implemented as a three-layer MLP (Multi-layer Perceptron) with Dropout. The input is the concatenated features of two linked (macro-)nodes ($2\times64=128-$D), and the output is the 1-D edge score. Experiments show that the edge discriminator helps to increase the discrimination of body-keypoint relational features, leading to better grouping accuracy. 
	
	\subsubsection{Macro-node Discriminator.} 
	We propose the macro-node discriminator to directly supervise the grouping procedure. After each grouping iteration, the nodes are clustered into macro-nodes. We apply a \emph{shared} macro-node discriminator to each macro-node to classify whether all keypoint candidates in the group belong to the same person (label 1) or not (label 0). Both the final human-level grouping results and the intermediate part-level grouping results are supervised. This provides denser supervision signals, facilitating the model training. The discriminator takes the aggregated macro-node features (64-D) as input and forwards it into a three-layer MLP to discriminate positive vs negative macro-nodes.
	
	\subsection{Implementation Details}
	
	\subsubsection{Keypoint Proposal Network.}
	The keypoint proposal network generates both 2D Gaussian confidence heatmaps as well as the pairwise relational feature maps. 2D Gaussian confidence heatmaps~\cite{cao2017realtime,newell2017associative,nie2017generative} are used to encode the keypoint locations and the ground truth confidence map for an image is calculated as the maximum of every person. We follow~\cite{cao2017realtime,newell2017associative} to apply keypoint NMS and parse the heatmaps to generate keypoint candidates. The pairwise relational feature maps are learned with push/pull losses, by pushing features of different people apart and pulling together features extracted from the same person.

	\subsubsection{Training and Inference.} 
	We implement OHGC based on AE~\cite{newell2017associative}\footnote{https://github.com/princeton-vl/pose-ae-train}. The input size is set as $512 \times 512$ and the output size is $128 \times 128$. The keypoint proposal network is first pre-trained and the keypoint proposal network, GNN and the edge/macro-node discriminators are jointly trained in an end-to-end manner. The losses include keypoint detection loss, pairwise pull/push losses, binary cross-entropy (BCE) loss for discriminators. The weights to balance these losses are set as $1:1e^{-3}:1e^{-5}$. We use Adam with an initial learning rate $2e^{-4}$ to train the model. During inference, flip testing and multi-scale testing is adopted. Unlike previous methods~\cite{cao2017realtime,newell2017associative}, we do not use single-person refinement.

	\section{Experiments}
	\label{sec:exp}
	
	\subsection{Datasets and Evaluation}
	To verify the effectiveness of the proposed HGG, we compare it with state-of-the-art methods on two challenging datasets, i.e. MS-COCO~\cite{lin2014microsoft}, and OCHuman~\cite{zhang2019pose2seg}. We follow~\cite{andriluka2017posetrack} to use Average Percision~(AP) to evaluate the methods.
	
	\textbf{MS-COCO Dataset}~\cite{lin2014microsoft} contains over 200,000 images and 250,000 human instances and 1.7 million labeled keypoints in total, among which 150,000 instances are for training and 80,000 instances are for testing. Our models are trained on the train set only. The ablation studies are reported on the val set and the comparisons with other state-of-the-arts are reported on the test-dev.

	\textbf{OCHuman Dataset}~\cite{zhang2019pose2seg} is a recently proposed benchmark to examine the limitations of human pose detection in highly challenging scenarios, which does not contain training samples and is intended to be used for evaluating existing models. It consists of 4731 images for validation and 8110 images for testing. The dataset contains only challenging cases of occlusion and the average IoU of the bounding boxes is 67\%. Following \cite{zhang2019pose2seg}, we train models on the training set of MS-COCO, and report the AP of them.

	\begin{table}[t]
		\centering
		\caption{(a) Comparisons with both top-down and bottom-up methods on COCO2017 test-dev dataset. $^{*}$ means using single-person pose refinement. $^{\times}$ means using extra segmentation annotation. $^{+}$ means using multi-scale test. Not that our results are obtained without single-person pose refinement.(b) Comparisons with both top-down and bottom-up methods on OCHuman dataset. Our results are obtained without single-person pose refinement.} 
		\begin{subtable}[b]{0.45\textwidth}
			\centering
			\tiny
			\begin{tabular}{c|c|c|c|c|c|c}
				\hline  
				Method &  $\operatorname{AP}$ & $\operatorname{AP}^{50}$ & $\operatorname{AP}^{75}$ & $\operatorname{AP}^{M}$ & $\operatorname{AP}^{L}$ & $\operatorname{AR}$\\
				\hline
				\multicolumn{7}{l}{\emph{Top-down methods}}\\
				\hline
				Mask-RCNN~\cite{he2017mask}& $63.1$ & $87.3$&$68.7$&$57.8$&$71.4$&$-$\\
				G-RMI~\cite{papandreou2017towards} &$64.9$ & $85.5$&$71.3$&$62.3$&$70.0$&$69.7$\\
				IPR~\cite{sun2018integral}  &$67.8$ & $88.2$&$74.8$&$63.9$&$74.0$&$-$\\
				CPN~\cite{chen2018cascaded}  & $72.1$ & $91.4$&$80.0$&$68.7$&$77.2$&$78.5$\\
				RMPE~\cite{fang2017rmpe}  &$72.3$ & $89.2$&$79.1$&$68.0$&$78.6$&$-$\\
				CFN~\cite{huang2017coarse}  & $72.6$ & $86.1$&$69.7$&$78.3$&$64.1$&$-$\\
				SBL~\cite{xiao2018simple} &${73.7}$ & ${91.9}$&${81.1}$&${70.3}$&${80.0}$&${79.0}$\\
				HRNet-W$48$~\cite{sun2019deep} & ${75.5}$&${92.5}$&${83.3}$&${71.9}$&${81.5}$&${80.5}$\\ \hline
				\multicolumn{7}{l}{\emph{Bottom-up methods}}\\
				\hline
				OpenPose$^{*}$~\cite{cao2017realtime} &$61.8$ & $84.9$&$67.5$&$57.1$&$68.2$&$66.5$\\
				AE$^{*+}$~\cite{newell2017associative}  &$65.5$ & $86.8$&$72.3$&$60.6$&$72.6$&$70.2$\\
				PersonLab$^{+\times}$~\cite{papandreou2018personlab}  &$68.7$ & $89.0$&$75.4$&$64.1$&$75.5$&$75.4$\\
				Directpose$^{+}$~\cite{tian2019directpose} & $64.8$ & $87.8$ &$71.1$ & $60.4$&$71.5$&$-$\\
				SPM$^{*+}$~\cite{nie2019single}  &$66.9$ & $88.5$&$72.9$&$62.6$&$73.1$&$-$\\
				Ours$^{+}$ &$67.6$&$85.1$&$73.7$&$62.7$&$74.6$&$71.3$\\ \hline \hline
			\end{tabular}
			
			\caption{}
			\label{tab:coco_test_dev_all}
		\end{subtable}
		\quad
		\begin{subtable}[b]{0.5\textwidth}
			\centering
			\scriptsize
			\setlength{\tabcolsep}{2.0pt}
			\begin{tabular}{l|c|c|c}
				\hline
				OCHuman & Backbone &  Val & Test  \\
				\hline
				\multicolumn{4}{l}{\emph{Top-down methods}} \\ \hline
				RMPE~\cite{fang2017rmpe} & Hourglass &${38.8}$ &${30.7}$  \\
				SBL~\cite{xiao2018simple} & ResNet50 &${37.8}$ & ${30.4}$ \\
				SBL~\cite{xiao2018simple} & ResNet152 &${41.0}$ & ${33.3}$ \\
				\hline
				\multicolumn{4}{l}{\emph{Bottom-up methods}} \\ \hline
				AE~\cite{newell2017associative} & Hourglass  &$32.1$ & $29.5$\\
				AE$^{+}$~\cite{newell2017associative} & Hourglass &$40.0$ & $32.8$\\
				Ours &  Hourglass &$35.6$ & $34.8$\\
				Ours$^{+}$  & Hourglass  & \textbf{41.8} & \textbf{36.0}\\ \hline
			\end{tabular}
			\caption{}
			\label{tab:OChuman}
		\end{subtable}
	\end{table}

	\subsection{Ablation Study}
	We validate the effectiveness of key modules in HGG by conducting the following ablation studies. For fair comparisons, all models use Hourglass as the backbone network and are trained with the same data augmentation and training schedule.
	
	\begin{table*}[t]
		\centering
		\scriptsize
		\caption{Ablation study of HGG's components on the COCO validation dataset.``FinalM'' means the final level macro-node discriminator. ``Edge'' means edge discriminator. `` IntermM'' means intermediate macro-node discriminator. ``MS'' means multi-scale testing. }
		\begin{tabular}{c|c|c|c|c|c|c|c|c|c|c|c|c}
			\hline
			\# & Method & Clustering  & FinalM & Edge & IntermM. & MS & $\operatorname{AP}$ & $\operatorname{AP}^{50}$ & $\operatorname{AP}^{75}$  & $\operatorname{AR}$ & $\operatorname{AR}^{50}$ & $\operatorname{AR}^{75}$ \\
			\hline
			1 & AE~\cite{newell2017associative} &  & & & & &$57.6$ &$79.7$ & $62.6$ & $62.1$ & $81.4$ & $66.1$ \\
			2 & AE~\cite{newell2017associative} &  & & & & \checkmark &$65.6$ &$85.1$ & $71.9$ & $69.1$ & $86.7$ & $74.2$ \\ \hline
			3 & Ours &\checkmark  & \checkmark& &  &  & $58.2$ & $80.8$ & $63.9$ & $63.4$ & $83.5$ & $68.0$ \\
			4 & Ours &\checkmark  &  & \checkmark &  &  & $59.6$ & $81.3$ & $65.1$ & $64.2$ & $83.0$ & $69.0$ \\
			5 & Ours &\checkmark  & \checkmark& \checkmark & & & $60.1$ & $81.6$ & $66.0$ & $64.5$ & $83.4$ & $69.6$ \\
			6 & Ours &  & \checkmark&\checkmark &\checkmark & & $59.6$ & $81.9$ & $65.5$  & $63.9$ & $83.3$ & $68.4$ \\
			7 & Ours-FC &\checkmark  & \checkmark&\checkmark &\checkmark & & $58.3$ & $80.7$ & $63.3$  & $62.5$ & $82.1$ & $66.9$ \\
			8 & Ours-GAT &\checkmark  & \checkmark&\checkmark &\checkmark & & $59.3$ & $81.1$ & $65.5$  & $63.9$ & $82.8$ & $69.0$ \\
			9 & Ours &\checkmark  & \checkmark&\checkmark &\checkmark & & $60.4$ & $83.0$ & $66.2$  & $64.8$ & $84.0$ & $69.8$ \\
			10& Ours &\checkmark  & \checkmark&\checkmark &\checkmark &\checkmark & $68.3$ & $86.7$ & $75.8$ & $72.0$ & $88.3$ & $78.0$ \\ \hline
		\end{tabular}
		\label{tab:ablation}
	\end{table*}
	
	\textbf{Effectiveness of End-to-End Learning.}
	We compare the performance of the baseline Associate Embedding (AE) model and that with the grouping loss. The grouping loss is provided by the final level macro-node discriminator. As shown in Table~\ref{tab:ablation} \#1 and \#3, end-to-end learning can increase the AP and the AR of the baseline by 0.6\% and 1.3\% respectively. \#6 uses all these grouping losses to train the models, but uses original post-processing greedy grouping during inference. The improvement of \#6 over \#1 indicates that the grouping loss and end-to-end learning can improve the capability of Keypoint Proposal Network. Note that under this setting, the grouping module can be removed during inference without adding any additional computation overhead.
	
	\textbf{Effectiveness of the Edge Discriminator.}
	The edge discriminator can enhance the keypoint relational features, thereby improving the grouping accuracy.
	To verify this, we compare the performance of models with and without the edge discriminator. As shown in Table~\ref{tab:ablation} \#1 and \#4, we find that supervising the linkage of the edge will significantly improve the grouping performance by $2.0$ mAP, demonstrating the effectiveness of the edge discriminator.
	
	\textbf{Effectiveness of the Macro-Node Discriminator.}
	We evaluate two kinds of macro-node supervision, intermediate macro-node supervision and final macro-node supervision. As shown in \#4 and \#5, the final macro-node supervision improves the grouping performance by $0.5$ mAP. By performing intermediate supervision to the macro-node, the result is further improved by $0.3$ mAP, shown in \#5 and \#9. In total, the full supervision boosts the performance by $0.8$ mAP, showing the importance of supervising the whole grouping process.

	\textbf{Effectiveness of GNN.}
	To evaluate the interaction GNN, we add two baselines for comparison. Ours-GAT uses GAT~\cite{velivckovic2018graph}, a popular graph neural network, for replacing EdgeConv. Ours-FC uses the multi-layer perception (dubbed FC for fully connected layers). For fair comparisons, these models have approximately the same parameter counts. As shown in \#7, \#8 and \#9, both graph-based models perform better than Ours-FC baseline, because of more effective interactive message passing. Moreover, EdgeConv (60.4 AP) performs the best.

\begin{figure}[t]
		\centering
		\begin{subfigure}[b]{0.5\textwidth}
			\begin{minipage}[t]{1\textwidth}
				\centering
				\includegraphics[width=0.95\linewidth]{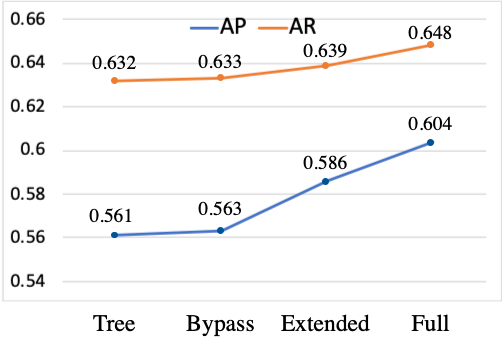}
				\caption{}
				\label{fig:graph_config}
			\end{minipage}
		\end{subfigure}
		\begin{subfigure}[b]{0.33\textwidth}
			\begin{minipage}[t]{1\textwidth}
				\centering
				\includegraphics[width=0.95\linewidth]{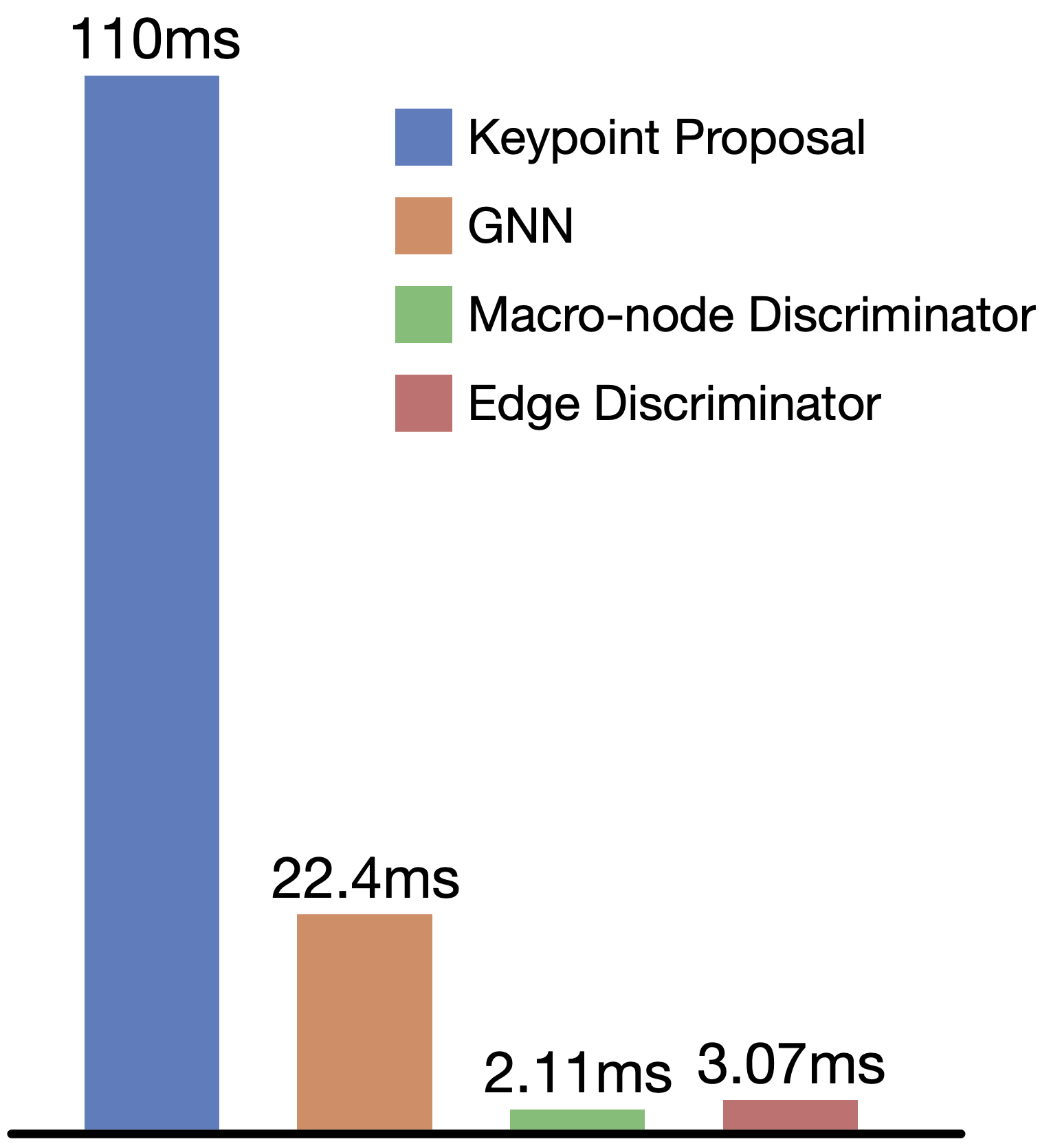}
				\caption{}
				\label{fig:time}
			\end{minipage}
		\end{subfigure}
		\caption{(a) Comparisons of different graph configurations on the COCO val set. Fully-connected graph (Full) performs the best among them. (b) Runtime analysis measured on one GTX-1060 GPU. The grouping module is very efficient compared to the keypoint proposal module.}
	\end{figure}

	\textbf{Comparisons of Different Graph Configurations.}
	As shown in Fig~\ref{fig:graph_config}, four types of commonly used graph configurations~\cite{doering2018joint} (\ie Tree, Bypass, Extended and Full) are compared. From Tree (the standard tree-structured model) to Full (the fully-connected graph), the graph gets denser. Bypass and Extended model adds some skip connections to the standard tree-structured model. As the complexity of the graph (or the number of connections) increases (Tree-Bypass-Extended-Full), the grouping accuracy increases from 56.1\% to 60.4\% mAP. In addition, the runtime of different graph configurations is almost the same. Therefore, we choose the fully-connected graph in our implementation.
	
	\subsection{Qualitative Analysis}
\begin{figure*}[t]
		\begin{center}
			\includegraphics[width=0.78\linewidth]{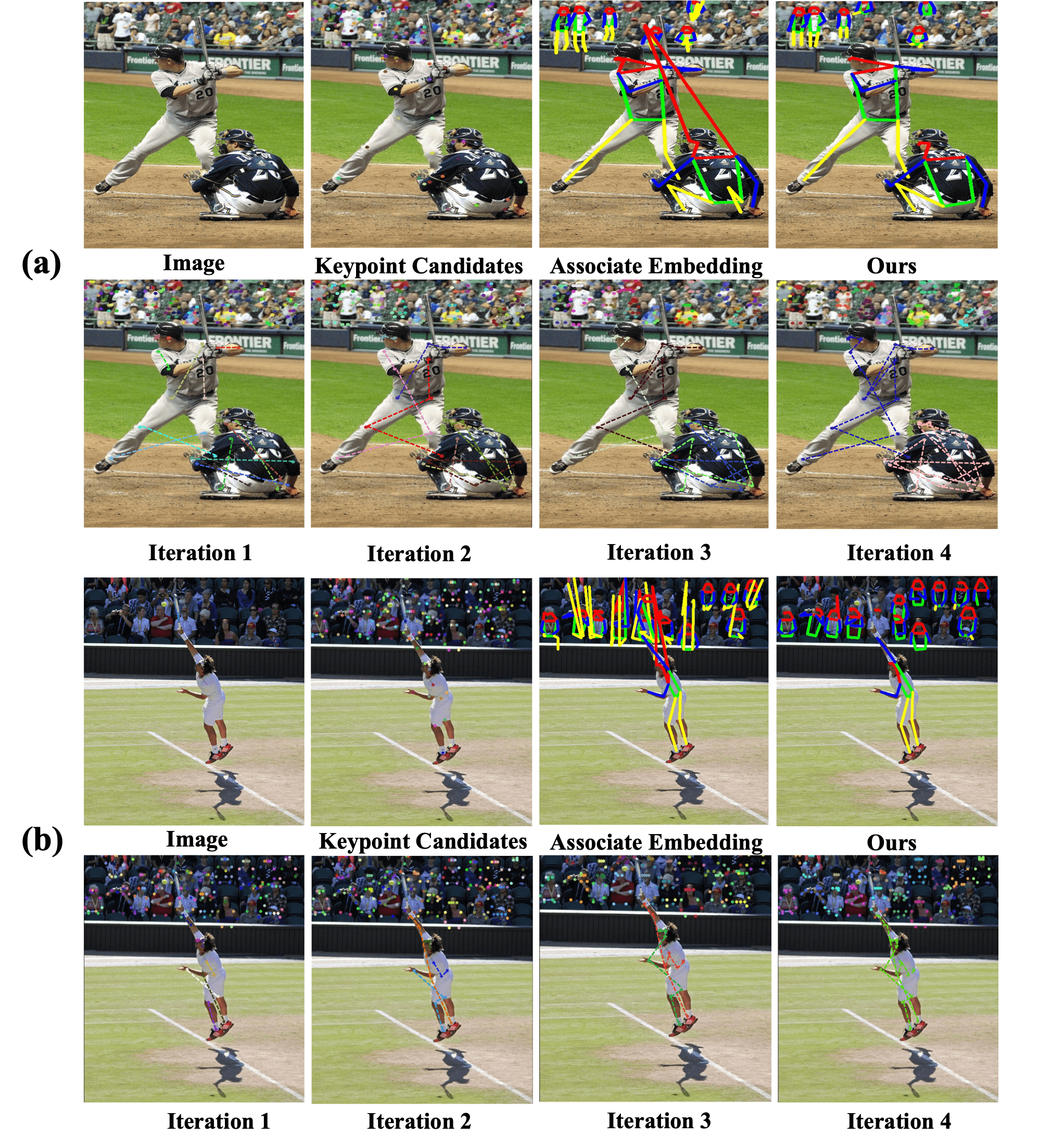}
		\end{center}
		\caption{The grouping process visualization. We show the grouped keypoint clusters in each iteration. Different colors are used to indicate different clusters. }
		\label{fig:analysis}
	\end{figure*}

	In Figure~\ref{fig:analysis}, we visualize the grouping procedure of OHGC algorithm. We use different colors to denote different clusters and dashed lines to highlight the macro-node merging process. OHGC starts with a set of keypoint candidates, each of which belongs to its own cluster. The grouping is performed iteratively. In each iteration, the most easy-to-group keypoints are merged. We show that the grouping procedure forms a pose hierarchy, from part to whole. Our method benefits from global supervision, which helps improve the grouping performance. 
	
	For failure cases, however, the current model is not able to recover false negatives or localization errors. Tiny people in images can lead to false negatives. Severe occlusion and non-typical poses may lead to localization errors. More test-time augmentation such as multi-scale testing, may mitigate these issues.
	
	\subsection{Comparisons with the State-of-the-art Methods}
	We compare our framework with the state-of-the-art methods on two large-scale multi-person pose estimation benchmarks.
	
	\textbf{Results on MSCOCO dataset}
	Table~\ref{tab:coco_test_dev_all} shows experimental results on MSCOCO $\tt test$-$\tt dev$ set. We see that the proposed HGG model achieves overall 67.6 AP. which is slightly lower than the state-of-the-art method PersonLab~\cite{papandreou2018personlab}. However, PersonLab uses extra annotations for instance segmentation. 
	Moreover, we also compare our method with recent single-shot methods (SPM~\cite{nie2019single} and DirectPose~\cite{tian2019directpose}). Surprisingly, although ours are lower than them in AP$^{50}$, in AP$^{75}$ ours are superior to them. This further indicates that our methods have advantages in scenarios that require high-precision pose estimation.
	
	\textbf{Results on OCHuman Dataset}
	To verify the robustness of HGG and other methods, we evaluate the proposed HGG model on the more challenging OCHuman dataset. We can see that our method achieves 41.8\% and 36.0\% mAP on val and test set, establishing a new state-of-the-art. Especially, HGG even outperforms top-down method SBL with 2.7 AP in test set, which further indicates our method is robust on more challenging scenarios.

	\subsection{Runtime Analysis}
	We analyze the time cost of the modules in HGG. Specifically, we evaluate our method on val set of MS-COCO and calculate the average time cost per image as shown in Fig.~\ref{fig:time}. The results are tested using PyTorch with a batchsize of 1 on one GTX-1060 GPU in a single thread. We find that the time cost of the grouping module is only a small proportion of the total time cost.
	
	\section{Conclusion and Future Work}
	\label{sec:conclusion}
	
	In this paper, we have reformulated the human pose estimation problem using the graph model and presented a full end-to-end learning framework named HGG. We have shown how we can combine the representative feature learning ability of CNN and the efficient long-range message passing as well as the relational feature learning capability of GNN. The macro-node discriminator and the edge discriminator are introduced to supervise the whole grouping process. We envision that the proposed framework can also be applied to other related problems such as multi-object tracking and instance segmentation. We expect to see more research in this direction in the near future.
	
	~\\
	\textbf{Acknowledgement.} This work is partially supported by the SenseTime Donation for Research, HKU Seed Fund for Basic Research, Startup Fund, General Research Fund No.27208720, the Australian Research Council Grant DP200103223 and Australian Medical Research Future Fund MRFAI000085.
	
	%
	%
	\bibliographystyle{splncs04}
	\bibliography{egbib}
\end{document}